\documentclass[11pt]{article}
\usepackage{authblk}
\usepackage{acl2016}
\usepackage{times}
\usepackage[hyphens]{url}
\usepackage{latexsym}
\usepackage{comment}
\usepackage{mathtools}
\pdfoutput=1

\aclfinalcopy 

\title{Reordering rules for English-Hindi SMT}
\author{\bf Raj Nath Patel}
\author{\bf Rohit Gupta}
\author{\bf Prakash B. Pimpale}
\author{\bf Sasikumar M}
\affil{CDAC Mumbai, Gulmohar Cross Road No. 9, Juhu, Mumbai-400049 India}
\affil{\{rajnathp,rohitg,prakash,sasi\}@cdac.in}

\date{}

\begin{document}
\maketitle
\begin{abstract}
Reordering is a preprocessing stage for Statistical Machine Translation (SMT) system where the words of the source sentence are reordered as per the syntax of the target language. We are proposing a rich set of rules for better reordering. The idea is to facilitate the training process by better alignments and parallel phrase extraction for a phrase-based SMT system. Reordering also helps the decoding process and hence improving the machine translation quality. We have observed significant improvements in the translation quality by using our approach over the baseline SMT. We have used BLEU, NIST, multi-reference word error rate, multi-reference position independent error rate for judging the improvements. We have exploited open source SMT toolkit MOSES to develop the system.
\end{abstract}


\section{Introduction} 
\label{sec:intro}
This paper describes syntactic reordering rules to reorder English sentences as per the Hindi language structure. Generally, in reordering approach, the source sentence is parsed (E) and syntactic reordering rules are applied to form reordered sentence (E`). The training of SMT system is performed using a parallel corpus, having source side reordered (E`) and target side. The decoding is done by supplying reordered source sentences. The source sentences prior to decoding are reordered using the same syntactic rules as applied to the training data. So, this process works as a preprocessing stage for the phrase-based SMT system. It has been observed that reordering as a preprocessing stage is beneficial for developing English-Hindi phrase-based SMT system~\cite{ramanathan:2008,rama:2014}. This paper describes a rich set of rules for the structural transformation of English sentence in Hindi language structure using Stanford~\cite{de:2006} parse tree on the source side. These rules are manually extracted based on analysis of source sentence tree and Hindi translation. 

For the evaluation purpose, we have trained and evaluated three different phrase-based SMT systems using MOSES toolkit~\cite{koehn:2007} and GIZA++~\cite{och:2003}. The first system was non-reordered baseline~\cite{brown:1990,marcu:2002,koehn:2003}, second using limited reordering described in~\newcite{ramanathan:2008} and third using improved reordering technique proposed in the paper. Evaluation has been carried out for end to end English-Hindi translation outputs using BLEU score~\cite{papineni:2002}, NIST score~\cite{doddington:2002}, multi-reference position-independent word error rate~\cite{tillmann:1997}, multi-reference word error rate~\cite{niessen:2000}. We have observed improvement in each of these evaluation metrics used. Next section discusses related work. Section~\ref{sec:approach} describes our reordering approach followed by experiments and results in section~\ref{sec:results} and conclusion in section~\ref{sec:conclusion}.


\section{Related Work}
\label{sec:bg}
Various pre-processing approaches have been proposed for handling syntax within SMT systems. These proposed methods reconcile the word-order differences between the source and target language sentences by reordering the source prior to the SMT training and decoding stages. For English-Hindi statistical machine translation, prior reordering is used by~\newcite{ramanathan:2008} and~\newcite{rama:2014}. This approach~\cite{ramanathan:2008} has shown significant improvements over baseline~\cite{brown:1990,marcu:2002,koehn:2003}. With the reordering, the BLEU score for the system has increased from 12.10 to 16.90. The same reordering approach~\cite{ramanathan:2008} used by us (with the different test set), has shown a slight improvement in BLEU score of 0.64 over baseline i.e. BLEU score increased from 21.55 to 22.19 compare to +4.8 BLEU point increase in the previous case. The reason can be, when the system is able to get bigger chunks from the phrase table itself the local reordering (within the phrase) is not needed and the long distance reordering employed in the earlier approach will be helpful for overall better translation. It may not be able to show significant improvements when local reordering is not captured by the translation model. 

Other language pairs have also shown significant improvement when reordering is employed. \newcite{xia:2004} have observed improvement for French-English and~\newcite{wang:2007} for Chinese-English language pairs. \newcite{niessen:2004} have proposed sentence restructuring whereas~\newcite{collins:2005} have proposed clause restructuring to improve German-English SMT.~\newcite{popovic2:2006} have also reported the use of simple local transformation rules for Spanish-English and Serbian-English translation. Recently, \newcite{khalilov:2011} proposed a reordering technique using a deterministic approach for long distance reordering and non-deterministic approach for short distance reordering exploiting morphological information. Some reordering approaches are also presented exploiting the SMT itself~\cite{gupta:2012,dlougach:2013}. Various evaluation techniques are available for reordering and overall machine translation evaluation. Particularly for reordering,~\newcite{birch:2010} have proposed LRScore, a language independent metric for evaluating the lexical and word reordering quality. The translation evaluation metrics include BLEU~\cite{papineni:2002}, Meteor~\cite{lavie:2009}, NIST~\cite{doddington:2002}, etc.


\section{Reordering approach}
\label{sec:approach}
Our reordering approach is based on a syntactic transformation of the English sentence parse tree according to the target language (Hindi) structure. It is similar to~\newcite{ramanathan:2008} but the transformation rules are not restricted to “SVO to SOV” and “pre-modifier to postmodifier” transformations only.

The idea was to come up with generic syntactic transformation rules to match the target language grammar structure. The motivation came from the fact that if words are already in a correct place with respect to other words in the sentence, the system doesn’t need to do the extra work of reordering at the decoding time. This problem becomes even more complicated when the system is not able to get bigger phrases for translating a sentence. Assuming an 18-word sentence, if the system is able to get only 2-word length phrases, there are 362880 (9!) translations (permutations) possible (still ignoring the case where one phrase having more than one translation options) for a sentence.

The source and the target sentences are manually analyzed to derive the tree transformation rules. From the generated set of rules, we have selected rules which seemed to be more generic. There are cases where we have found more than one possible correct transformations for an English sentence as the target language (Hindi) is a free word order language within certain limits. In such cases, word order close to English structure is preferred over other possible word orders with respect to Hindi. 

We identified 5 categories which are most prominent candidates for reordering. These include VPs (verb phrases), NPs (noun phrases), ADJPs (adjective phrase), PPs (preposition phrase) and ADVPs (adverb phrase). In the following subsections, we have described rules for these in more detail.

\begin{table}[!hbt]
	\centering
	\begin{tabular}{|l|p{5.5cm}|}
		\hline
		\multicolumn{1}{|l|}{\bf Tag}  & \multicolumn{1}{c|}{\bf Description (Penn tags)} \\ \hline
		\textit{dcP} & \textit{Any, parser generated phrase} \\ \hline
		\textit{pp} & \textit{Prepositional Phrase (PP)} \\ \hline
		\textit{whP} & \textit{WH Phrase (WHNP, WHADVP, WHADJP, WHPP)} \\ \hline
		\textit{vp} & \textit{Verb Phrase (VP)} \\ \hline
		\textit{sbar} & \textit{Subordinate clause (SBAR)} \\ \hline
		\textit{np} & \textit{Noun phrase (NP)} \\ \hline
		\textit{vpw} & \textit{Verb words (VBN, VBP, VB, VBG, MD, VBZ, VBD)} \\ \hline
		\textit{prep} & \textit{Preposition words (IN, TO, VBN, VBG)} \\ \hline
		\textit{adv} & \textit{Adverbial words (RB, RBR, RBS)} \\ \hline
		\textit{adj} & \textit{Adjunct word (JJ,JJR,JJS)} \\ \hline
		\textit{advP} & \textit{Adverb phrase (ADVP)} \\ \hline
		\textit{punct} & \textit{Punctuation (,)} \\ \hline
		\textit{adjP} & \textit{Adjective phrase (ADJP)} \\ \hline
		\textit{OP} & \textit{advP, np and/or pp} \\ \hline
		\textit{$Tag^*$} & \textit{One or more occurrences of Tag} \\ \hline
		\textit{Tag?} & \textit{Zero or one occurrence of Tag} \\ \hline
	\end{tabular}
	\caption{Tag description}
	\label{tab:tag-set}
\end{table}

The format for writing the rules is as follows: $Type\_of\_phrase\ (tag1\ tag2\ tag\ 3\ :\ tag2\ tag1\ tag3)$. This means that “$tag1\ tag2\ tag3$”, structure has been transformed to “$tag2\ tag1\ tag3$” for the $type\_of\_phrase$. This $type\_of\_phrase$ denotes our category (NP, VP, ADJP, ADVP, PP) in which rule falls. The Table~\ref{tab:tag-set} explains about various tags and corresponding Penn-tags used in the writing these rules. 

The following subsections explain the reordering rules. The higher precedence rule is written prior to the lower precedence. In general, the more specific rules have higher precedence. Each rule is followed by an example with intermediate steps of parsing and transformation as per the Hindi sentence structure. “Partial Reordered” shows the effect of the particular rule, whereas “Reordered” shows the impact of the whole reordering approach. The Hindi (transliterated) sentence is also provided as a reference to the corresponding English sentence.


\subsection{Noun Phrase Rules} \label{subsec:noun}
\begin{equation}
\begin{aligned}
NP\ (np1\ PP[\ prep\ NP[\ np2\ sbar]]\ : \\
\ np2\ prep\ np1\ sbar)
\end{aligned}
\end{equation}

English: \textit{The time of the year} when nature dawns all its colorful splendor, is beautiful.

Parse: \textit{[NP (np1 the time) [PP (prep of) [NP (np2 the year)} (sbar when nature dawns all its colorful splendor)]]] , is beautiful .

Partial Reordered: \textit{(np2 the year) (prep of) (np1 the time)} (sbar when nature dawns all its colorful splendor) , is beautiful . 

Reordered: \textit{(np2 the year) (prep of) (np1 the time)} (sbar when nature all its colorful splendor dawns) , beautiful is .

Hindi: varsh ka samay jab prakriti apne sabhi rang-birange vabahv failati hai, sundar hai.
\begin{equation}
\begin{aligned}
NP(np\ SBAR[\ S[\ dcP\ ]]\ :dcP\ np)
\end{aligned}
\end{equation}

English: September to march is \textit{the best season to visit Udaipur}.

Parse: September to March is  \textit{[NP (np the best season) [SBAR [S (dcP to visit Udaipur)]]]} .

Partial Reordered: September to March is  \textit{(dcP to visit Udaipur) (np the best season)} .

Reordered: September to March  \textit{(dcP Udaipur visit to) (np the best season) is} .

Hindi: september se march udaipur ghumane ka sabse achcha samay hai.
\begin{equation}
\begin{aligned}
NP(np\ punct\ advP\ :\ advP\ punct\ np)
\end{aligned}
\end{equation}

English: The modern town of \textit{Mumbai, about 50 km south of Navi Mumbai} is Kharghar.

Parse: The modern town of \textit{[NP (np Mumbai) (punct ,) (advP about 50 km south of Navi Mumbai)]} is Kharghar .

Partial Reordered: \textit{(advP about 50 km south of Navi Mumbai)) (punct ,) (dcP The modern town of Mumbai)} is kharghar .

Reordered: \textit{(advP Navi Mumbai of about 50 km south) (punct ,) (dcP Mumbai of the modern town)} kharghar is .

Hindi: navi mumbai ke 50 km dakshin me mumbai ka adhunic sahar kharghar hai.
\begin{equation}
\begin{aligned}
NP(\ np\ vp\ :\ vp\ np)
\end{aligned}
\end{equation}

English: The main attraction is \textit{a divine tree called as 'Kalptaru'}.

Parse: The main attraction is  \textit{[NP (np a divine tree) (vp called as 'Kalptaru') ]} .

Partial Reordered: The main attraction is  \textit{(vp` called as 'Kalptaru' ) (np a divine tree)} .

Reordered: The main attraction  \textit{(vp ` Kalptaru' as called) (np a divine tree) is} .

Hindi: iska mukhya akarshan kalptaru namak ek divya vriksh hai.

\subsection{Verb Phrase Rules}
\begin{equation}
\begin{aligned}
VP(\ vpw\ PP\ [\ prep\ NP[\ np\ punct?\ SBAR \\
[whP\ cP\ ]]]\ : np\ prep\ vpw\ punct?\ whP\ dcP)
\end{aligned}
\end{equation}

English: The best time to visit \textit{is in the afternoon} when the crowd thins out.

Parse: The best time to visit \textit{[VP (vpw is) PP[(prep in) NP[ (np the afternoon)} [SBAR (whP when) (dcP the crowd thins out)]]] .

Partial Reordered: The best time to visit \textit{(np the afternoon) (prep in) (vpw is)} (whP when) (dcP the crowd thins out) .

Reordered: visit to The best time \textit{(np the afternoon) (prep in) (vpw is)} (whP when) (dcP the crowd thins out) .

Hindi: bhraman karane ka sabase achcha samay dopahar me hai jab bhid kam ho jati hai.
\begin{equation}
\begin{aligned}
VP(\ vpw\ NP[\ np\ punct?\ SBAR[\ whP \\
dcP\ ]]\ :\ np\ vpw\ punct?\ whP\ dcP)
\end{aligned}
\end{equation}

English: Jaswant Thada \textit{is a white marble monument} which was built in 1899 in the memory of Maharaja Jaswant Singh II.

Parse: jaswant thada \textit{[VP (vpw is) [NP (np a white marble monument)} [SBAR (whP which) (dcP was built in 1899 in the memory of Maharaja Jaswant Singh II)]] .

Partial Reordered: Jaswant Thada \textit{(np a white marble monument) (vpw is)} (whP which) (dcP was built in 1899 in the memory of Maharaja Jaswant Singh II) .

Reordered: Jaswant Thada \textit{(np a white marble monument) (vpw is)} (whP which) (dcP Maharaja Jaswant Singh II of the memory in 1899 in built was) .

Hindi: jaswant thada ek safed sangmarmar ka smarak hai jo ki maharaja jaswant singh dwitiya ki yad me 1889 me banwaya gaya tha.
\begin{equation}
\begin{aligned}
VP(vpw\ OP\ sbar\ :\ OP\ vpw\ sbar)
\end{aligned}
\end{equation}

English: Temples in Bhubaneshwar are \textit{built beautifully on a common plan} as prescribed by Hindu norms.

Parse: Temples in Bhubaneshwar are \textit{[VP (vpw built) (advP beautifully) (pp on a common plan)} (sbar as prescribed by Hindu norms)] .
Partial Reordered: Bhubaneshwar in Temples are \textit{(advP beautifully) (pp a common plan on)} (vpw built) (sbar as prescribed by Hindu norms) .

Reordered: Bhubaneshwar in Temples \textit{(advP beautifully) (pp a common plan on)} (vpw built) are (sbar as Hindu norms by prescribed) .

Hindi: bhubaneswar ke mandir hindu niyamon dwara nirdharit samanya yojana ke anusar banaye gaye hain.
\begin{equation}
\begin{aligned}
VP(vpw\ pp1\ pp2^*\ :\ pp2^*\ pp1\ vpw)
\end{aligned}
\end{equation}

English: Avalanche is \textit{located at a distance of 28 Kms from Ooty}.

Parse: Avalanche is \textit{[VP (vpw located) (pp1 at a distance of 28 kms) (pp2 from Ooty)]} .

Partial Reordered: Avalanche is \textit{(pp2 from Ooty) (pp1 at a distance of 28 kms) (vpw located)}.

Reordered: Avalanche \textit{(pp2 Ooty from ) (pp1 28 kms of a distance at) (vpw located) is} .

Hindi: avalanche ooty se 28 km ki duri par sthit hai.
\begin{equation}
\begin{aligned}
VP(vpw\ np\ pp\ :\ np\ pp\ vbw)
\end{aligned}
\end{equation}

English: Taxis and city buses available outside the station, \textit{facilitate access to the city}.

Parse: Taxis and city buses available outside the station , \textit{[VP (vpw facilitate) (np access) (pp to the city)]} .

Partial Reordered: Taxis and city buses available outside the station , \textit{(pp to the city) (np access) (vpw facilitate)} .

Reordered: Taxis and city buses the station outside available , \textit{(pp the city to) (np access) (vpw facilitate)} .

Hindi: station ke baahar sahar jane ke liye taksi aur bus ki suvidha upalabdha hai.
\begin{equation}
\begin{aligned}
VP(\ prep\ dcP\ :\ dcP\ prep)
\end{aligned}
\end{equation}

English: A wall was built \textit{to protect it}.

Parse: A wall was built \textit{[VP (prep to) (dcP protect it)]} .

Partial Reordered: A wall was built \textit{(protect it) (prep to)} .

Reordered: A wall \textit{(dcP it protect) (prep to)} built was .

Hindi: ek diwar ise surakshit karane ke liye banayi gayi thi.
\begin{equation}
\begin{aligned}
VP(adv\ vpw\ dcphrase\ :\ dcphrase\ adv\ vpw)
\end{aligned}
\end{equation}

English: Modern artist such as French sculptor Bartholdi is \textit{best known by his famous work}.

Parse: Modern artists such as French sculptor Bartholdi is \textit{[VP (adv best) (vpw known) (dcP by his famous work)]} .

Partial Reordered: Modern artists such as French sculptor Bartholdi is \textit{(dcP by his famous work) (adv best) (vpw known)} .

Reordered: such as French sculptor Bartholdi Modern artists \textit{(dcP his famous work by) (adv best) (vpw known)} is .

Hindi: french shilpkar bartholdi jaise aadhunik kalakar apane prashidha kam ke liye vishesh rup se jane jate hain.
\begin{equation}
\begin{aligned}
VP(advP\ vpw\ dcP\ :\ advP\ dcP\ vpw)
\end{aligned}
\end{equation}

English: Bikaner, popularly known as the camel county is located in Rajasthan.

Parse: Bikaner , [VP (advP popularly) (vpw known) (dcP as the camel country)] is located in Rajsthan .

Partial Reordered: Bikaner , (advP popularly) (dcP as the camel country) (vpw known) is located in Rajsthan .

Reordered: Bikaner , (advP popularly) (dcP the camel country as) (vpw known) Rajsthan in located is .

Hindi: bikaner , jo aam taur par unton ke desh ke naam se jana jata hai, rajasthan me sthit hai.
\begin{equation}
\begin{aligned}
VP(vpw\ adv?\ adjP?\ dcP\ :\\
dcP\ adjP?\ adv?\ vpw)
\end{aligned}
\end{equation}

English: This palace has \textit{been beautiful from many years}.

Parse: This palace has \textit{[VP (vpw been) (adjP beautiful) (dcP from many years)]} .

Partial Reordered: This palace has \textit{(dcP from many years) (adjP beautiful) (vpw been)} .

Reordered: This palace \textit{(dcP many years from) (adjP beautiful) (vpw been) has}.

Hindi: yah mahal kai varson se sunder raha hai.
\subsection{Adjective and Adverb Phrase Rules}
\begin{equation}
\begin{aligned}
ADJP(\ vpw\ pp\ :\ pp\ vpw)
\end{aligned}
\end{equation}

English: The temple is \textit{decorated with paintings depicting incidents}.

Parse: The temple is \textit{[ADJP (vpw decorated) (pp with paintings depicting incidents )]} .

Partial Reordered: The temple is \textit{(pp with paintings depicting incidents) (vpw decorated)} .

Reordered: The temple \textit{(pp incidents depicting paintings with) (vpw decorated)} is .

Hindi: mandir ghatnao ko darshate hue chitron se sajaya gya hai.
\begin{equation}
\begin{aligned}
ADJP(\ adjP\ pp\ :\ pp\ adjP\ )
\end{aligned}
\end{equation}

English: As a resul, temperatures are now \textit{higher than ever} before.

Parse: As a result , temperatures are now \textit{[ADJP (adjP higher) (pp than ever)]} before .

Partial Reordered: As a result , temperatures are now \textit{(pp than ever) (adj higher)} before .

Reordered: a result As , temperatures now before \textit{(pp ever than) (adj higher)} are .

Hindi: parinam swarup taapman ab pahle se bhi adhik hai.
\begin{equation}
\begin{aligned}
ADJP(\ adj\ dcP\ :\ dcP\ adj\ )
\end{aligned}
\end{equation}

English: The Kanha National park is \textit{open to visitors}.

Parse: The Kanha National park is \textit{[ADJP (adj open) (dcP to visitors)]} .

Partial Reordered: The Kanha National park is \textit{(pp to visitors ) (adj open)} .

Reordered: The Kanha National park \textit{(pp visitors to) (adj open)} is .

Hindi: kanha national park paryatakon ke liye khula hai.
\begin{equation}
\begin{aligned}
ADVP(\ adv\ dcP\ :\ dcP\ adv\ )
\end{aligned}
\end{equation}

English: The temple is most favored spot for tourists \textit{apart from the pilgrims}.

Parse: The temple is most favored spot for tourists \textit{[ADVP (adv apart) (dcP from the pilgrims)]} .

Partial Reordered: The temple is most favored spot for tourists \textit{(dcP from the pilgrims ) (adv apart)} .

Reordered: The temple most favored spot \textit{(dcP the pilgrims from) (adv apart) is} .

Hindi: mandir teerth yatriyon ke alawa paryatkon ke liye bhi lokpriya sthal hai.

\subsection{Preposition Phrase Rules}
\begin{equation}
\begin{aligned}
PP(\ adv\ prep?\ dcP\ :\ dcP\ prep?\ adv\ )
\end{aligned}
\end{equation}

English: Does kalajar occur \textit{because of sun}?

Parse: Does kalajar occur \textit{[PP (adv because) (prep? of) (dcP sun)]} ?

Partial Reordered: Does kalajar occur \textit{(dcp sun) (prep? of) (adv because)} ?

Reordered: Does kalajar \textit{(dcp sun) (prep? of) (adv because)} occur ?

Hindi: kya kalajar dhup ke karan hota hai?


\section{Experiments and Results}
\label{sec:results}
The experiments were carried out on the corpus described in Table~\ref{tab:data} below.

\begin{table}[!hbt] \small
	\centering
	\begin{tabular}{|l|c|c|}
		\hline
		\multicolumn{1}{|l|}{} & \multicolumn{1}{c|}{\#sentences} & \multicolumn{1}{c|}{\#words} \\ \hline
		\multicolumn{1}{|l|}{training} & 94926 & 1235163 \\ \hline
		\multicolumn{1}{|l|}{tuning} & 1446 & 23600 \\ \hline
		\multicolumn{1}{|l|}{testing} & 500 & 9792 \\ \hline
	\end{tabular}
	\caption{Corpus distribution}
	\label{tab:data}
\end{table}

\begin{table}[!hbt]\small
	\centering
	\begin{tabular}{|p{1.3cm}|c|c|c|c|}
		\hline
		& BLEU & NIST & mWER & mPER \\ \hline
		baseline & 21.55 & 5.72 & 68.08 & 45.54 \\ \hline
		limited RO & 22.19 & 5.74 & 66.44 & 44.70 \\ \hline
		our approach & 24.47 & 5.88 & 64.71 & 43.89 \\ \hline
	\end{tabular}
	\caption{Evaluation scores; RO: Reordering}
	\label{tab:results}
\end{table}

\begin{table*}[!hbt] \small
	\centering
	\begin{tabular}{|p{1.7cm}|p{12.5cm}|}
		\hline
		input & Ahmedabad was named after the sultan Ahmed Shah, who built the city in 1411.(en) \\ \hline
		baseline & ahmedabad was named after the sultan ahmed shah, who built the city in 1411.(en) \\
		& \textit{ahamadaabaada ke naama para rakhaa gayaa sultaana ahamada shaaha vaale shahara 1411} (hi) \\ \hline
		limited RO & ahmedabad the sultan ahmed shah , who the city 1411 in built after named was .(en) \\
		& \textit{ahamadaabaada kaa naama sultaana ahamadashaaha ke , jisane 1411 meM shahara banavaayaa ke naama para rakhaa gayaa thaa.}(hi) \\ \hline 
		our approach & ahmedabad the sultan ahmed shah after named was , who 1411 in the city built .(en) \\
		& \textit{ahamadaabaada kaa naama sultaana ahamadashaaha ke naama se paDaa thaa jisane 1411 meM shahara banavaayaa thaa.}(hi) \\ \hline
		reference & \textit{ahamadaabaada kaa naama sultaana ahamadashaaha ke naama para paDaa thaa , jisane 1411 meM shahara banavaayaa thaa.}(hi) \\ \hline
	\end{tabular}
	\caption{Comparison of translation on a sentence from test corpus}
	\label{tab:example}
\end{table*}

\begin{table*}[!hbt] \small
	\centering
	\begin{tabular}{|p{1cm}|p{1cm}|p{2.5cm}|p{2cm}|p{1cm}|p{2.5cm}|p{2cm}|}
		\hline
		\multicolumn{1}{|p{1cm}|}{} & \multicolumn{3}{|p{6cm}|}{\#phrases} & \multicolumn{3}{|p{6cm}|}{\#distinct-phrases(distinct on source)} \\ \hline
		\multicolumn{1}{|p{1cm}|}{phrase length} & \multicolumn{1}{|p{1cm}}{baseline} & \multicolumn{1}{|p{2.5cm}|}{limited RO /\%IOBL/IOBL} & \multicolumn{1}{|p{2cm}}{our approach /\%IOBL/IOBL} & \multicolumn{1}{|p{1cm}}{baseline} & \multicolumn{1}{|p{2.5cm}}{limited RO /\%IOBL/IOBL} & \multicolumn{1}{|p{2cm}|}{our approach /\%IOBL/IOBL} \\ \hline
		2 & 537017 & 579878/ 7.98/ 42861 & 579630/ 9.98/ 42613 & 208988 & 249847/ 19.55/ 40859 & 254393/ 21.72/ 45405 \\ \hline
		3 & 504810 & 590265/ 16.92/	85455 & 616381/ 22.10/ 111571 & 292183 & 384518/ 31.62/ 92335 & 408240/ 39.72/ 116057 \\ \hline
		4 & 406069 & 493637/ 21.56/	87568 & 531904/ 30.98/ 125835 & 268431 & 372282/ 38.68/ 103851 & 409966/ 52.72/ 141535 \\ \hline
		5 & 313368 & 391490/ 24.92/ 78122 & 431135/ 37.58/ 117766 & 221228 & 313723/ 41.80/ 92495 & 354273/ 60.13/ 133045 \\ \hline
		6 & 231146 & 292899/ 26.71/	61753 & 327192/ 41.55/ 96046 & 170852 & 244643/ 43.19/ 73791 & 279723/ 63.72/ 108871 \\ \hline
		7 & 154800 & 196679/ 27.05/	41879 & 220868/ 42.67/ 66068 & 119628 & 170108/ 42.19/ 50480 & 194881/ 62.90/ 75253 \\ \hline
	\end{tabular}
	\caption{Phrase count analysis}
	\label{tab:stat}
\end{table*}

The baseline system was setup by using the phrase-based model~\cite{brown:1990,marcu:2002,koehn:2003}. For the language model, we carried out experiments and found in a comparison that 5-gram model with modified Kneser-Ney smoothing~\cite{chen:1996} to be the best performing. Target Hindi corpus from the training set was used for creating the language model. The KenLM~\cite{heafield:2011} toolkit was used for the language modeling experiments. The tuning corpus was used to set weights for the language models, distortion model, phrase translation model, etc., using minimum error rate training~\cite{och:2003}. Decoding was performed using the MOSES decoder. Stanford constituency parser~\cite{de:2006} was used for parsing. 

Table~\ref{tab:example} describes with the help of an \footnotemark example \footnotetext{All hindi words have been written in Itrans using  \url{http://sanskritlibrary.org/transcodeText.html}} how the reordering and hence the translation quality has improved. From the example, it can be seen that the translation of the system using our approach is better than the other two systems. The output translation is structurally more correct in our approach and conveys the same meaning with respect to the reference translation.

The Table~\ref{tab:results} lists four different evaluations of the systems under study. For BLEU and NIST higher score is considered as better and for mWER and the mPER lower score is desirable. Table~\ref{tab:results} shows the results of comparative evaluation of baseline, limited reordering and our approach with improved reordering. We find that the addition of more reordering rules shows substantial improvements over the baseline phrase-based system and the limited reordering system~\cite{ramanathan:2008}. The impact of improved syntactic reordering can be seen as the BLEU and NIST scores have increased, whereas mWER and mPER scores have decreased.

Table~\ref{tab:stat} shows the count of overall phrases and distinct phrases (distinct on the source) for baseline, limited reordering approach, and our improved reordering approach. The table also shows an increase over baseline (IOBL) and percentage increase over baseline(\%IOBL) for limited reordering and improved reordering. We have observed that no. of distinct phrases extracted from the training corpus get increased. The \%IOBL for bigger phrases is more compared to shorter phrases. This can be attributed to the better alignments resulting in the extraction of more phrases~\cite{koehn:2003}.

We have also observed that the overall increase is even lesser than the increase in no. of distinct phrases (distinct on the source) for all the phrase-lengths in our approach (e.g. 42613 and 45405 for phrase-length 2) which shows that reordering makes word alignments more consistent and reduces multiple entries for the same source phrase. The training was done on maximum phrase length 7(default).

\section{Conclusion}
\label{sec:conclusion}
It can be seen that the addition of more reordering rules improves translation quality. As of now, we have tried these rules only for English-Hindi pair, but the plan is to employ similar reordering rules in other English-Indian language pairs as most Indian languages are structurally similar to Hindi. Also, plans are there going for a comparative study of improved reordering system and hierarchical models.

\bibliography{acl2016}
\bibliographystyle{acl2016}

\end{document}